\pdfoutput=1

\documentclass[11pt]{article}

\usepackage[final]{acl}

\usepackage{times}
\usepackage{latexsym}
\usepackage[T2A,T1]{fontenc}
\usepackage[utf8]{inputenc}
\usepackage[russian,english]{babel}
\usepackage{microtype}
\usepackage{inconsolata}
\usepackage{xspace}
\usepackage{multirow}
\usepackage{booktabs}
\usepackage{bm}
\usepackage{amsmath}
\usepackage{amsfonts}
\usepackage{hyperref}
\usepackage{graphicx}
\usepackage{framed}
\usepackage{enumitem}
\usepackage{pifont}
\usepackage{CJKutf8}
\usepackage{marvosym}

\newcommand{\ie}{\emph{i.e.,}\xspace}

\newcommand{\eg}{\emph{e.g.,}\xspace}
\newcommand{\wrt}{\emph{w.r.t.}\xspace}

\newcommand{\etc}{\emph{etc}}
\newcommand{\ignore}[1]{}

\makeatletter
\def\@fnsymbol#1{}
\makeatother

\title{Not All Metrics Are Guilty: \\ Improving NLG Evaluation by Diversifying References}

\author{
Tianyi Tang\textsuperscript{\rm{1,6 *}}\thanks{*\ This work was done during internship at MSRA.}, 
Hongyuan Lu\textsuperscript{\rm{4}}, 
Yuchen Eleanor Jiang\textsuperscript{\rm{5}}, 
Haoyang Huang\textsuperscript{\rm{2}}, \\
\textbf{Dongdong Zhang}\textsuperscript{\rm{2}}\textbf{,} 
\textbf{Wayne Xin Zhao}\textsuperscript{\rm{1,6 \Letter}\thanks{\textsuperscript{\Letter}\ Corresponding author}\ }\textbf{,}
\textbf{Tom Kocmi}\textsuperscript{\rm{3}}\textbf{,}
\textbf{Furu Wei}\textsuperscript{\rm{2}} \\
\textsuperscript{1} Gaoling School of Artificial Intelligence, Renmin University of China \\
\textsuperscript{2} Microsoft Research Asia, China \quad
\textsuperscript{3} Microsoft \\
\textsuperscript{4} The Chinese University of Hong Kong \quad \textsuperscript{5} AIWaves Inc. \\
\textsuperscript{6} Beijing Key Laboratory of Big Data Management and Analysis Methods \\
\texttt{\{steventianyitang,hongyuanlu\}@outlook.com \ \ eleanor.jiang@aiwaves.cn} \\
\texttt{\{haohua,dozhang,tomkocmi,fuwei\}@microsoft.com \ \ \ batmanfly@gmail.com} \\
} 

\begin{document}
\maketitle
\begin{abstract}
Most research about natural language generation~(NLG) relies on evaluation benchmarks with limited references for a sample, which may result in poor correlations with human judgements. The underlying reason is that one semantic meaning can actually be expressed in different forms, and the evaluation with a single or few references may not accurately reflect the quality of the model's hypotheses. 
To address this issue, this paper presents a simple and effective method, named \textbf{Div-Ref}, to enhance existing evaluation benchmarks by enriching the number of references. We leverage large language models~(LLMs) to diversify the expression of a single reference into multiple high-quality ones to cover the semantic space of the reference sentence as much as possible. 
We conduct comprehensive experiments to empirically demonstrate that diversifying the expression of reference can significantly enhance the correlation between automatic evaluation and human evaluation. This idea is compatible with recent LLM-based evaluation which can similarly derive advantages from incorporating multiple references.
\textit{We strongly encourage future generation benchmarks to include more references, even if they are generated by LLMs, which is once for all.}
We release all the code and data at \url{https://github.com/RUCAIBox/Div-Ref} to facilitate research.
\end{abstract}

\section{Introduction}
Evaluation plays a pivotal role in advancing the research on natural language generation~(NLG)~\cite{metric_survey2,tg_survey}. 
It aims to measure the quality of the generated hypotheses in NLG tasks (\eg machine translation, text summarization, and image caption) from multiple aspects, such as accuracy, fluency, informativeness, and semantic consistency. 
There exist two typical approaches for NLG evaluation, namely human evaluation and automatic evaluation. Human evaluation relies on qualified annotators for a reliable assessment of the generation results of NLG models~\cite{metric_survey}.  
However, it is very costly and time-consuming to conduct large-scale human evaluations, especially for complicated tasks.

\begin{table}[t]
    \centering
    \small
    \resizebox{\columnwidth}{!}{
    \begin{tabular}{l|l}
        \toprule
        Input $\mathbf{x}$ & \begin{CJK*}{UTF8}{gbsn}苹果是我最喜欢的水果，但香蕉是她的最爱。\end{CJK*} \\
        \midrule
        Reference $\mathbf{y^*}$ & The apple is my most loved fruit but the banana is her most loved. \\
        \midrule
        Hypothesis $\hat{\mathbf{y}}$ & My favorite fruit is apple, while hers beloved is banana. \\
        \midrule
        \multicolumn{2}{l}{\textbf{BLEU$(\hat{\mathbf{y}}$|$\mathbf{y^*})=0.014$, \qquad \qquad \qquad  BERTScore$(\hat{\mathbf{y}}$|$\mathbf{y^*})=0.923$}} \\ 
        \midrule
        \multirow{3}{*}{\begin{tabular}[c]{@{}c@{}}Diversified \\ references \\ $\tilde{\mathbf{y}}_1,\tilde{\mathbf{y}}_2,\tilde{\mathbf{y}}_3$ \end{tabular}}
        & Apples rank as my favorite fruit, but bananas hold that title for her. \\
        & Apple is my favorite fruit, but banana is her most beloved. \\
        & My most loved fruit is the apple, while her most loved is the banana. \\
        \midrule
        \multicolumn{2}{l}{\textbf{BLEU$(\hat{\mathbf{y}}$|$\mathbf{y^*},\tilde{\mathbf{y}}_1,\tilde{\mathbf{y}}_2,\tilde{\mathbf{y}}_3)=0.251$, \quad BERTScore$(\hat{\mathbf{y}}$|$\mathbf{y^*},\tilde{\mathbf{y}}_1,\tilde{\mathbf{y}}_2,\tilde{\mathbf{y}}_3)=0.958$}}      \\
        \bottomrule
    \end{tabular}}
    \caption{The motivation illustration of our proposed Div-Ref method. For the Chinese-to-English translation, the evaluation scores of BLEU and BERTScore are relatively low when using the single ground-truth reference. After diversify the ground truth into multiple references, the correlation of these two metrics with human evaluation can be improved.}
    \label{tab:example}
\end{table}


To reduce the human cost, researchers have proposed various automatic evaluation metrics. 
Yet, due to their rigid analytic forms,  they often suffer from an inaccurate approximation of the task goal, even having significant discrepancies with human evaluation~\cite{zhang2023benchmarking}.  
Despite the widespread concerns about evaluation metrics~\citep{sulem-etal-2018-bleu,alva-manchego-etal-2021-un}, another seldom discussed yet important factor is the \emph{number of reference} texts in the evaluation benchmarks. 
There always exist diverse hypotheses that would satisfy the goal of an NLG task, however, the number of ground-truth references provided by human annotators is often limited in scale. For example, there is only one English ground-truth reference written for a Chinese input sentence in the WMT22 News Translation Task~\cite{wmt22}. 
This potentially leads to unreliable evaluation results when using limited ground-truth references, as illustrated in Table~\ref{tab:example}. 




Considering the above-mentioned issue, this paper attempts to improve the NLG evaluation benchmarks and make existing automatic metrics better reflect the actual quality of the hypotheses.
We focus on increasing the number of reference texts to narrow the gap between automatic and human evaluation. 
The key idea is to leverage the excellent ability of existing LLMs to provide more high-quality references for a single sample. By enriching the diversity of the references while maintaining semantic consistency, we expand the coverage of the semantic expressions for evaluating the generated texts from \emph{a single or few}  standard references to \emph{a more diverse set} of semantically equivalent references. In this way, our evaluation method can better approximate human evaluation criteria, as the improved scores shown in Table~\ref{tab:example}.
In addition, increasing the number of references is \emph{agnostic} to specific task settings and can be integrated with various automatic metrics for evaluating different generation tasks.  

To demonstrate the effectiveness of diversifying references, we conduct extensive experiments on the benchmarks from multiple NLG tasks. 
The experimental results demonstrate that incorporating multiple references can significantly improve the consistency between traditional evaluation metrics and human evaluation results. Surprisingly, it is even applicable in multilingual and multimodal text generation scenarios.
Importantly, our approach is orthogonal with automatic metrics, enabling even the recent LLM-based evaluations~\cite{gemba,wang2023chatgpt} to benefit from our diversified references and achieve SOTA correlation with human judges.
\textit{Therefore, incorporating more references for the NLG benchmark proves advantageous, requiring a one-time effort, and future researchers can reap its benefits.}

\section{Related Work}

\subsection{Automatic Evaluation}
Automatic evaluation metrics for natural language generation could be mainly categorized into two streams: reference-based and reference-free evaluation. The former involves measuring the quality of the hypothesis by comparing it with single or few ground-truth references, \eg BLEU~\cite{bleu}, ROUGE~\cite{rouge}, and METEOR~\cite{meteor}. They primarily focus on the n-gram overlaps between the hypothesis and the references. Recently, neural metrics have become a mainstream method to evaluate semantic similarity and usually have a higher correlation with human evaluation. The representative metrics include BERTScore~\cite{bertscore}, BLEURT~\cite{bleurt}, and recent methods involving LLMs~\cite{gemba, wang2023chatgpt,chiang2023can,luo2023chatgpt,lu2023error,gao2023human}.
Reference-free evaluations assess the hypothesis without the necessity of any reference. They often adopt neural-based models as a black box for evaluating semantic quality as well as grammatical fluency~\citep{zhao-etal-2020-limitations, mehri-eskenazi-2020-usr, hessel-etal-2021-clipscore, gpteval, chen2023exploring}. 
However, the reference-free metrics has lower correlation with human compared to the reference-based ones~\cite{gemba,wang2023chatgpt}. 
In this work, we primarily focus on the reference-based automatic metrics, even without the need for altering their core implementation.

\subsection{Increasing the Reference Number}
Initially, researchers attempt to utilize paraphrasing methods~\cite{bandel-etal-2022-quality} to enrich the instances of training set~\cite{zheng-etal-2018-multi,khayrallah-etal-2020-simulated}.
\citet{zhou-etal-2006-paraeval} use paraphrasing to enhance the evaluation of the summarization task. There are also prior works that employed paraphrasing in enhancing evaluations with machine translation, either by human paraphrasing~\citep{gupta-etal-2019-investigating,freitag-etal-2020-bleu,freitag-etal-2020-human} or automatic paraphrasing~\citep{zhou-etal-2006-evaluating,kauchak-barzilay-2006-paraphrasing,thompson-post-2020-automatic,bawden-etal-2020-study,bawden-etal-2020-parbleu}. 
One recent study reports that the maximization of diversity should be favored for paraphrasing~\cite{bawden-etal-2020-study}, which enhances the succeeding evaluation.
Although current work showcases the promise of paraphrasing methods, they are confined to improving the correlation of specific metrics (\eg BLEU and ROUGE) in certain tasks (\eg translation and summarization). They neglect to explore the importance of the number of references, considering constraints such as the quality of automatic paraphrasing or the expense of human paraphrasing.
Meanwhile, our investigation reveals that the majority of newly proposed NLG benchmarks in 2023 continue to rely on only one reference. Even those benchmarks incorporating multiple references typically feature no more than two or three ground truth.
The advent of LLMs has facilitated a convenient and effective means of diversifying references to encompass the semantic space of samples.
In this work, we design dedicated prompts tailored for LLMs and extensively investigate the imperative of augmenting the number of references in NLG benchmarks.

\section{Methodology}
This section first provides a formal definition by introducing several crucial aspects of NLG evaluation. We then describe our approach that leverages LLMs to enrich the semantic coverage of references, bridging the gap between automatic evaluation and human evaluation.

\subsection{NLG Evaluation Formulation}
As for an NLG task, let $\mathbf{x}$ denote the input sequence associated with extra information (task goal, additional context, \etc) and $\mathbf{y}^*$ denote the ground-truth reference provided by the benchmark.
After a model or system generates the hypothesis sequence $\hat{\mathbf{y}}$, the automatic evaluation of the metric $\mathcal{M}$ can be represented as $\mathcal{M}(\hat{\mathbf{y}}|\mathbf{x},\mathbf{y}^*)$.
Accordingly, we can also represent human evaluation as $\mathcal{H}(\hat{\mathbf{y}}|\mathbf{x},\mathbf{y}^*)$.  Hence, to access the quality of the metric $\mathcal{M}$, researchers usually calculate the correlation score with human evaluation $\mathcal{H}$:
\begin{equation}
\rho (\mathcal{M}(\hat{\mathbf{y}}|\mathbf{x},\mathbf{y}^*), \mathcal{H}(\hat{\mathbf{y}}|\mathbf{x},\mathbf{y}^*)),
\end{equation}
where $\rho$ can be any correlation function such as Spearman correlation and Kendall’s tau. An ideal metric is to maximize the correlation between automatic evaluation $\mathcal{M}$ and human evaluation $\mathcal{H}$.

Note that, $\mathcal{H}$ is a subjective process and cannot be directly calculated. Intuitively, when a human assesses on the hypothesis $\hat{\mathbf{y}}$, he or she will match $\hat{\mathbf{y}}$ among various valid sentences, which can be illustrated as a semantic sentence space $\mathbb{Y}$ formed in our brain based on human knowledge and common sense related to the ground-truth reference $\mathbf{y}^*$. Therefore, the human evaluation can be further described as $\mathcal{H}(\hat{\mathbf{y}}|\mathbf{x},\mathbb{Y})$.

While researchers on NLG evaluation focus on proposing various implementations of $\mathcal{M}$, we aim to improve the automatic evaluation benchmark using $\mathcal{M}(\hat{\mathbf{y}}|\mathbf{x},A(\mathbb{Y}))$, where $A(\mathbb{Y})$ is the approximation of $\mathbb{Y}$ to instantiate the semantic space. $A(\mathbb{Y})$ is defined as $\{\mathbf{y}^*,\tilde{\mathbf{y}}_1,\dots,\tilde{\mathbf{y}}_n\}$ to alleviate the bias and insufficiency of a single reference in representing the entire semantic space of the ground-truth references. To achieve this, we augment the reference with diverse expressions while retaining the same meaning, aiming to approximate the semantic space $\mathbb{Y}$. In the traditional single-reference evaluation benchmark, $A(\mathbb{Y})$ corresponds to $\{\mathbf{y}^*\}$.

As the acquisition of $A(\mathbb{Y})$ is costly for human annotation, we propose to leverage the superior capability of LLMs to generate high-quality and diverse references. With this approach, the automatic evaluation can be formulated as follows:
\begin{equation} 
\mathcal{M}(\hat{\mathbf{y}}|\mathbf{x},A(\mathbb{Y})) = \mathcal{M}(\hat{\mathbf{y}}|\mathbf{x},\mathbf{y}^*,\tilde{\mathbf{y}}_1,\dots,\tilde{\mathbf{y}}_n).
\end{equation}
Traditional metrics, such as BLEU~\cite{bleu} and ChrF~\cite{chrf}, have built-in algorithms to handle multiple references, while for neural metrics, they only support a single reference and then aggregate the scores from each reference.
In practice, the evaluation score under the multiple-reference setting can be calculated as follows:
\begin{equation} \label{eq:final}
\mathcal{M}(\hat{\mathbf{y}}|\mathbf{x},\mathbf{y}^*,\tilde{\mathbf{y}}_1,\dots,\tilde{\mathbf{y}}_n)=\mathop{\mathcal{F}}_{i=0}^n \big[\mathcal{M}(\hat{\mathbf{y}}|\mathbf{x},\hat{\mathbf{y}_i})\big],
\end{equation}
where $\hat{\mathbf{y}_0}=\mathbf{y}^*$ and $\mathcal{F}$ is a function leveraged to aggregate scores of multiple diversified sequences, which can be the operation of maximum aggregation or mean aggregation.


\subsection{LLM Diversifying for Evaluation} \label{sec:para}
Recently, LLMs have showcased remarkable capabilities across various NLP tasks. They have proven to be powerful aids in tasks such as text paraphrasing, text style transfer, and grammatical error correction~\cite{llm_paraphrase}. Therefore, we harness the potential of LLMs as the approximation function $A$ to generate diverse expressions $\tilde{\mathbf{y}}_1,\dots,\tilde{\mathbf{y}}_n$ while preserving the original semantics of the ground-truth reference $\mathbf{y}^*$.


\subsubsection{Paraphrasing Prompt}
Following existing work~\cite{bawden-etal-2020-study}, we provide the LLM with the paraphrasing prompt ``Paraphrase the sentences: \texttt{\{reference\}}'' to wrap the given reference and employ nucleus sampling~\cite{nucleus} to generate a variety of rephrased sentences. In our preliminary experiments, we apply the paraphrasing prompt to paraphrase ten sentences for each English reference sentence from the WMT22 Metrics Shared Task~\cite{wmt22_metric}. 
We calculate a semantic diversity score\footnote{We calculate the mean cosine distance between each rephrased pair using OpenAI Embeddings \texttt{text-embedding-ada-002}. Then, we average the score of each instance to obtain an overall semantic diversity score.} of the rephrased sentences as 0.032. We further observe that rephrased sentences primarily involve word-level substitutions, with minimal modifications to the sentence structure.

\subsubsection{Diversified Prompts}
To improve the diversity of the reference sentences as suggested by~\citet{bawden-etal-2020-study}, we explore several heuristic rules to obtain more diverse texts and cover the semantic space. Inspired by~\citet{jiao2023chatgpt}, we ask ChatGPT to provide instructions that cover different aspects of semantic expressions with the prompt: ``\textit{Provide ten prompts that can make you diversify the expression of given texts by considering different aspects.}''. According to the suggestions by~\citet{savage2006effective}, we screen out ten diversifying instructions to promote the changes in words, order, structure, voice, style, \etc, which are listed as follows:
\begin{framed}
\small

\noindent \ding{192} Change the order of the sentences:
\vspace{0.5em}

\noindent \ding{193} Change the structure of the sentences:
\vspace{0.5em}

\noindent \ding{194} Change the voice of the sentences:
\vspace{0.5em}

\noindent \ding{195} Change the tense of the sentences:
\vspace{0.5em}

\noindent \ding{196} Alter the tone of the sentences:
\vspace{0.5em}

\noindent \ding{197} Alter the style of the sentences:
\vspace{0.5em}

\noindent \ding{198} Rephrase the sentences while retaining the original meaning:
\vspace{0.5em}

\noindent \ding{199} Use synonyms or related words to express the sentences with the same meaning:
\vspace{0.5em}

\noindent \ding{200} Use more formal language to change the level of formality of the sentences:
\vspace{0.5em}

\noindent \ding{201} Use less formal language to change the level of formality of the sentences:
\end{framed}

Then, we also utilize the ten instructions to generate ten diversified sentences in total (\ie one for each instruction). The semantic diversity score increases from 0.032 to 0.049, which demonstrates a significant diversity improvement among the sentences and verifies the effectiveness of our diverse prompts. Note that, our diversifying method is not just paraphrasing but attempts to cover different aspects of the reference expressions. Considering the strong cross-lingual generation capabilities of LLMs~\cite{bloomz}, we apply English instructions to diversify references in different languages (\eg German and Russian). The diversified examples can be found in Tables~\ref{tab-zh}, \ref{tab-de}, \ref{tab-ru}.

\subsubsection{Discussion}
Compared with existing work~\cite{freitag-etal-2020-bleu,bawden-etal-2020-study} that utilizes paraphrasing for evaluation, we leverage the recent superior LLMs for diversifying the expressions of given reference. 
After supervised fine-tuning and reinforcement learning from human feedback, LLMs showcase excellent capability to follow the input instruction and align with human preference, which can not achieve by previous paraphrasing methods. 
To verify the effectiveness of LLMs, we further conduct experiments in Section~\ref{sec:ablation} to compare them with traditional paraphrasing models.
Moreover, we conduct experiments to evaluate the diversifying results of LLMs. We employ another excellent GPT 3.5 to judge whether the generated sentence conveys the same meaning of given reference. The results show that 94.6\% of the generated sentences are suitable, which demonstrates the effectiveness and robustness of our diverse prompts. Note that, LLM diversifying is simple and convenient and does not need any post manual filtering. We conduct further experiments to verify it in Section~\ref{sec:ablation}.

\section{Experiments}
In this section, we deliberately select three different types of natural language generation tasks to verify the effectiveness of multiple references.

\subsection{Experimental Setup}
\subsubsection{Benchmarks}
We choose three meta evaluation benchmarks covering multilingual and multimodal scenarios. These metric benchmarks consist of human scores of the generated text (\ie $\mathcal{H}(\hat{\mathbf{y}}|\mathbf{x},\mathbb{Y})$), and we can calculate their correlation with the automatic metric scores $\mathcal{M}(\hat{\mathbf{y}}|\mathbf{x},A(\mathbb{Y}))$ using multiple references.
\begin{itemize}[leftmargin=*]
\itemsep0em 
\item WMT22 Metrics Shared Task~\cite{wmt22_metric} includes the generated sentences of different competitor models in the WMT22 News Translation Task~\cite{wmt22}. They require human experts to rate these sentences via the multidimensional quality metrics (MQM) schema. We use all three evaluated language pairs, including Chinese (Zh)$\rightarrow$English (En), English (En)$\rightarrow$German (De), and English (En)$\rightarrow$Russian (Ru). 
We leverage the standardized toolkit \texttt{mt-metrics-eval V2}\footnote{\url{github.com/google-research/mt-metrics-eval}} to calculate the segment-level Kendall Tau score and the system-level pairwise accuracy following~\citet{kocmi-etal-2021-ship}. Note that the overall system-level pairwise accuracy across three languages is the most important metric for translation evaluation~\cite{deutsch2023ties}.

\item SummEval~\cite{summeval} comprises 200 summaries generated by each of the 16 models on the CNN/Daily Mail dataset~\cite{cnndm}. Human judgements measure these summaries in terms of coherence, consistency, fluency, and relevance. We apply the sample-level Spearman score to measure the correlation.

\item PASCAL-50S~\cite{cider} is a triple collection of 4,000 instances wherein each instance consists of one reference and two captions. Human annotators compare the two captions based on the reference and express their preference. We calculate the accuracy of whether the metric assigns a higher score to the caption preferred by humans. Our experiments follow the setups outlined by~\citet{hessel-etal-2021-clipscore}.
\end{itemize}

\subsubsection{Metrics}
We evaluate a variety of automatic metrics covering different categories. Based on the taxonomy of existing work~\cite{metric_survey}, we select 16 metrics subdivided into five classes:

\begin{itemize}[leftmargin=*]
\itemsep0em 
\item Character-based metrics: ChrF~\cite{chrf}; 

\item Word-based metrics: BLEU~\cite{bleu}, ROUGE-1/2/L~\cite{rouge}, METEOR~\cite{meteor}, CIDEr~\cite{cider}, and SPICE~\cite{spice};

\item Embedding-based metrics: BERTScore~\cite{bertscore} and MoverScore; 

\item Trained metrics: BLEURT~\cite{bleurt}, Prism~\cite{prism}, COMET~\cite{comet}, and BARTScore~\cite{bartscore};

\item LLM-based metrics: GEMBA-Dav3-DA~\cite{gemba} and ChatGPT-eval (Stars w/ ref)~\cite{wang2023chatgpt}; 
\end{itemize}

The implementation of each metrics are detailed Appendix~\ref{app:metric}. The metrics we used for each benchmark are listed in Table~\ref{tab:task-metric}.

\begin{table}[!t]
\centering
\resizebox{\columnwidth}{!}{
\begin{tabular}{c|l|ccc}
\toprule
\textbf{Categories} & \textbf{Metrics} & \textbf{Translation} & \textbf{Summarization} & \textbf{Caption} \\ 
\midrule
\textbf{Character} & ChrF & \checkmark & -- & -- \\
\midrule
\multirow{9.5}{*}{\textbf{Word}}
& BLEU & \checkmark & -- & \checkmark \\
\cmidrule{2-5}
& ROUGE-1 & -- & \checkmark & -- \\
\cmidrule{2-5}
& ROUGE-2 & -- & \checkmark & -- \\
\cmidrule{2-5}
& ROUGE-L & -- & \checkmark & \checkmark \\
\cmidrule{2-5}
& METEOR & -- & -- & \checkmark \\
\cmidrule{2-5}
& CIDEr & -- & -- & \checkmark \\
\cmidrule{2-5}
& SPICE & -- & -- & \checkmark \\
\midrule
\multirow{2.5}{*}{\textbf{Embedding}}
& BERTScore & \checkmark & \checkmark & \checkmark \\
\cmidrule{2-5}
& MoverScore & -- & \checkmark & -- \\
\midrule
\multirow{6}{*}{\textbf{Trained}}
& BLEURT & \checkmark & -- & -- \\
\cmidrule{2-5}
& Prism & \checkmark & -- & -- \\
\cmidrule{2-5}
& COMET & \checkmark & -- & -- \\
\cmidrule{2-5}
& BARTScore & \checkmark & -- & -- \\
\midrule
\multirow{2.5}{*}{\textbf{LLM}}
& GEMBA & \checkmark & -- & -- \\
\cmidrule{2-5}
& ChatGPT-eval & -- & \checkmark & -- \\
\bottomrule
\end{tabular}}
\caption{The summary of metrics evaluated on tasks.}
\label{tab:task-metric}
\end{table}

\begin{figure*}[!t]
    \centering
    \includegraphics[width=1.0\textwidth]{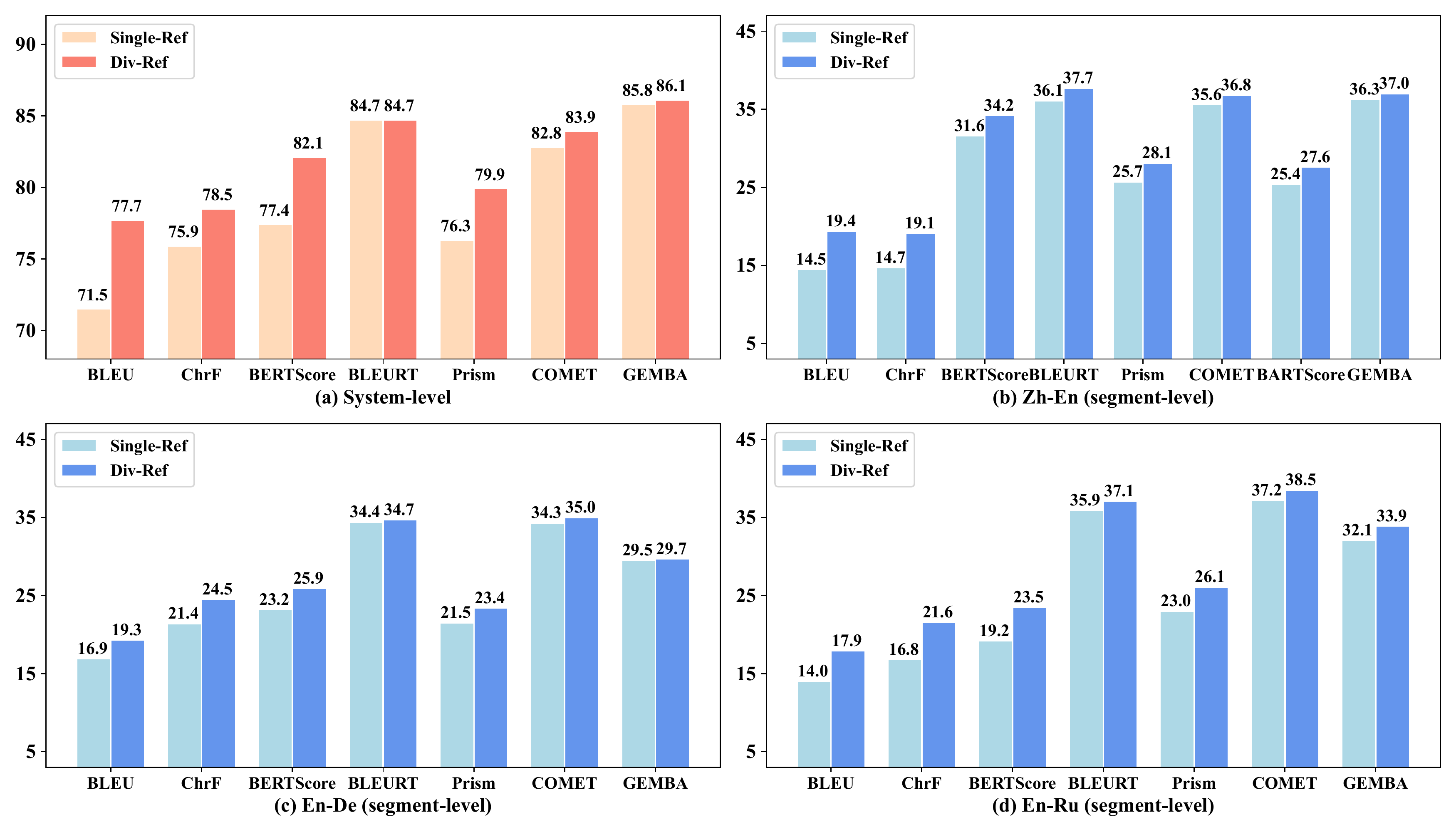}
    \caption{System-level pairwise accuracy (main aspect) and Kendall Tau correlation of segment-level score over the WMT22 Metrics Shared Task on three translation directions.}
    \label{fig:wmt}
\end{figure*}

\begin{figure*}[!t]
    \centering
    \includegraphics[width=1.0\textwidth]{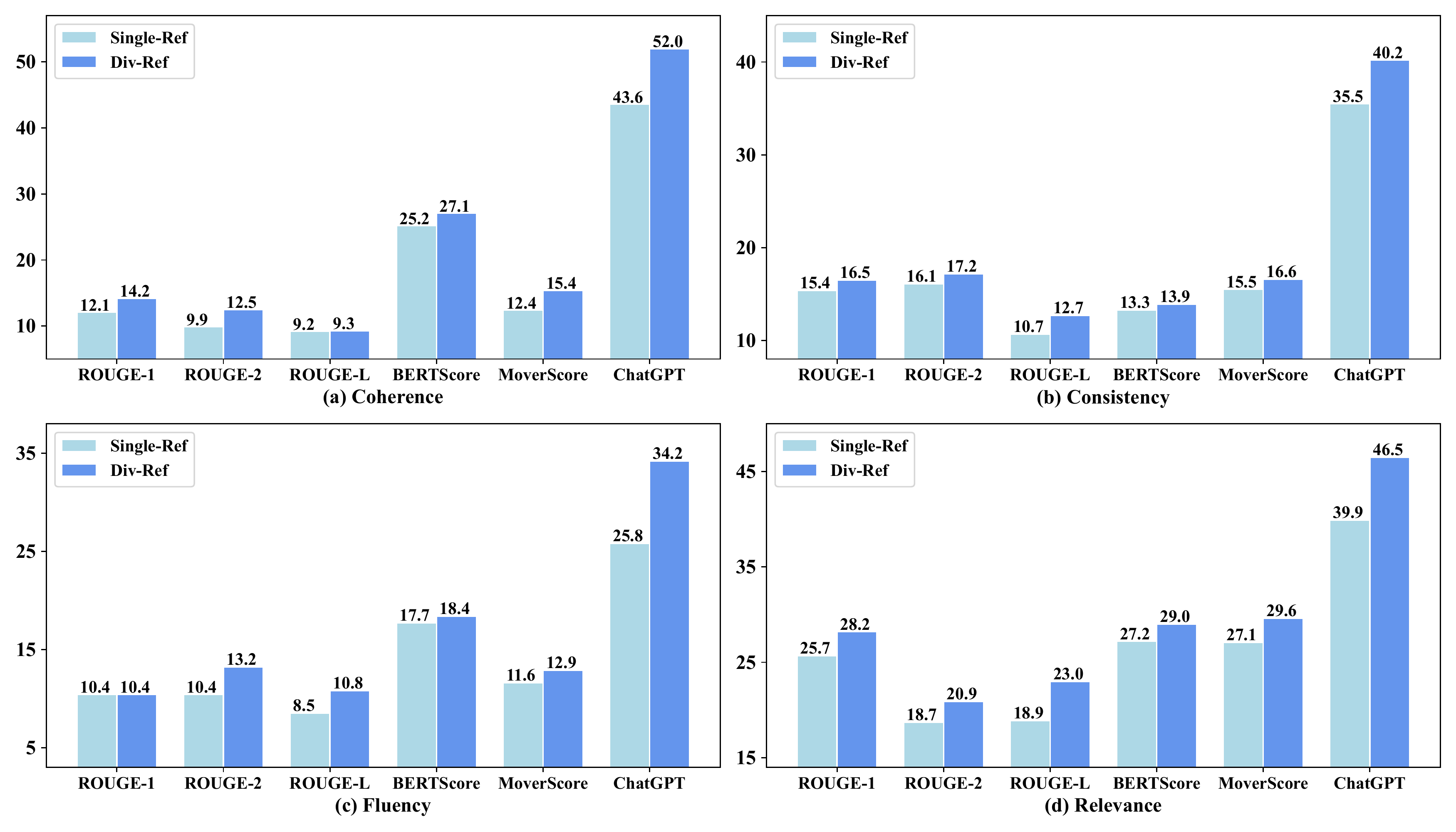}
    \caption{Spearman score of sample-level correlation over the SummEval benchmark on four evaluation aspects.}
    \label{fig:summeval}
\end{figure*}

\subsubsection{Implementation Details}
As for our approach, we utilize the \texttt{gpt-3.5-turbo-instruct} model as the LLM along with the instructions outlined in Section~\ref{sec:para} to diversify the reference sentences into different expressions. When utilizing the OpenAI API, we set the temperature to 1 and the top\_p to 0.9. In Equation~\ref{eq:final}, we employ the maximum aggregation and generate 10 diversified sentences (\ie one for each instruction). We further analyze these hyper-parameters in Section~\ref{sec:ablation}.

In our experiments, the baseline method is the evaluation of various metrics over single-reference benchmarks, represented by \textbf{Single-Ref}, and the evaluation of our approach over multiple diversified references is denoted as \textbf{Div-Ref}.

\subsection{Experimental Results}
The results of the three evaluation benchmarks over various automatic metrics are shown in the following subsections. We can see that enriching the number of references using our our LLM diversifying method shows a better correlation with human evaluation than the single-reference baseline.
Our method is also compatible with existing SOTA LLM-based methods and can enhance them to achieve a higher correlation.

\begin{figure*}[!t]
    \centering
    \includegraphics[width=1.0\textwidth]{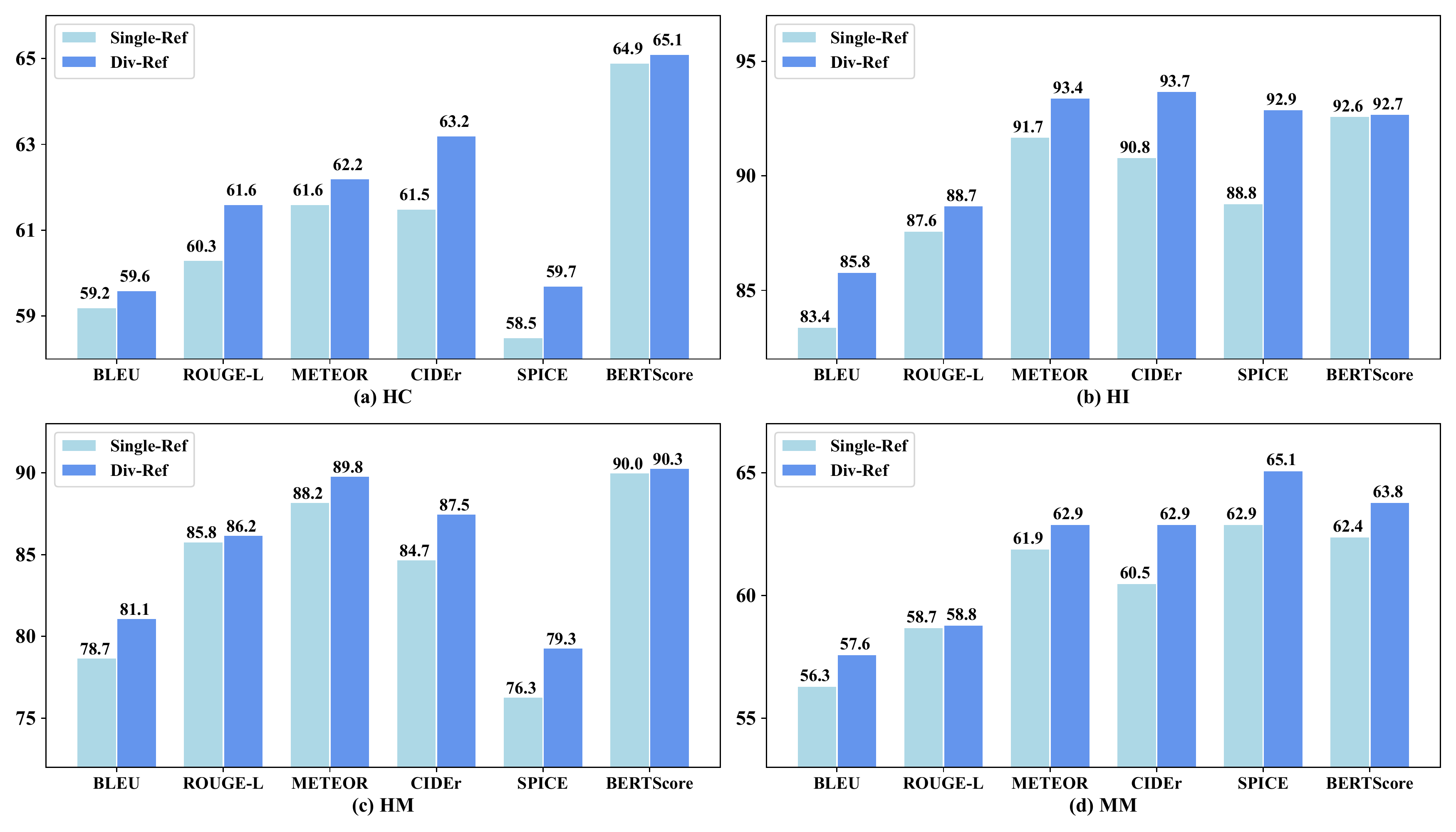}
    \caption{Accuracy score over the PASCAL-50S benchmark on four settings. HC denotes the two captions are correct and written by humans. HI denotes two human-written captions but one is irrelevant. HM denotes one caption is human-written and the other is model-generated. MM denotes two model-generated captions.}
    \label{fig:pascal}
\end{figure*}

\subsubsection{Evaluation on Machine Translation}
As shown in the Figure~\ref{fig:wmt}, our Div-Ref method has shown consistent correlation improvements across all evaluation when compared to the single-reference baseline.
Surprisingly, the SOTA metric GEMBA can still be enhanced when evaluated with more references.
In terms of different languages, we observe that the diversifying methods are effective across different languages. English and Russian references benefit more than the German ones, which may be due to the distinct multilingual ability of \texttt{gpt-3.5-turbo-instruct}.
Notably, our approach showcases significant effects on the traditional BLEU metric, which can further facilitate the application due to its efficiency and universality. The large improvement further demonstrates the automatic metric may be not guilty but the evaluation benchmark needs more references.

\subsubsection{Evaluation on Text Summarization}
In the summarization task, we select six metrics to examine the correlation against human evaluation from four aspects: coherence, consistency, fluency, and relevance. 
According to the results shown in Figure~\ref{fig:summeval}, the Div-Ref method can make significant improvements in almost all dimensions compared to the traditional single-reference approach. 
We can see that the traditional word-based metrics (\eg ROUGE) and the embedding-based metrics (\eg BERTScore) perform closely, while LLM-based metric shows remarkable correlation with human evaluation. 
This phenomena further demonstrates the effectiveness of LLMs for NLG evaluation, as described by~\citet{wang2023chatgpt}.
It should be noted that our method has further improved the LLM-based metric ChatGPT-eval in all dimensions. This also shows that our approach is effective in improving the correlation with human evaluation and the NLG benchmarks should include more references.

\subsubsection{Evaluation on Image Caption}
In order to examine the effectiveness of our method for the image caption task, we expand the reference under four different settings to judge whether the metric assigns a higher score to the caption preferred by humans. 
The results of the image caption task are reported in Figure~\ref{fig:pascal}.
For the HC and MM settings, which are difficult settings to judge two similar captions, Div-Ref exhibits enhancements in all metrics, particularly for SPICE, METEOR, and BERTScore. This verifies our approach can expand the semantic coverage of references to bridge the gap between automatic evaluation and human evaluation.
Regarding HI and HM, Div-Ref still maintains the improvements in all metrics, except for a slight drop for BERTScore in the HM setting. 
Despite one of the candidate captions being incorrect or machine-generated, our method can strongly align different metrics with human preference, particularly for the SPICE metric. In comparison to the single-reference baseline, our approach yields a significant improvement of 3.6 points with SPICE in HI and 2.9 points for HM.

\subsection{Ablation Analysis} \label{sec:ablation}
\begin{table*}[t]
    \centering
    \resizebox{1\textwidth}{!}{
    \begin{tabular}{ll|cc|cc|cc|cc|cc|cc|cc}
    \toprule
    \multicolumn{2}{c|}{\multirow{2}{*}{\textbf{Settings}}}                             & \multicolumn{2}{c|}{\textbf{BLEU}} & \multicolumn{2}{c|}{\textbf{ChrF}}  & \multicolumn{2}{c|}{\textbf{BERTScore}} & \multicolumn{2}{c|}{\textbf{BLEURT}} & \multicolumn{2}{c|}{\textbf{Prism}} & \multicolumn{2}{c|}{\textbf{COMET}} & \multicolumn{2}{c}{\textbf{Average Gains}} \\ 
    & & System & Zh-En & System & Zh-En & System & Zh-En & System & Zh-En & System & Zh-En & System & Zh-En & System & Zh-En \\
    \midrule
    \multicolumn{2}{c|}{\textbf{\textcolor{gray}{Single-Ref}}}                          & \textcolor{gray}{71.5} & \textcolor{gray}{14.5} & \textcolor{gray}{75.9} & \textcolor{gray}{14.7} & \textcolor{gray}{77.4} & \textcolor{gray}{31.6} & \textcolor{gray}{84.7} & \textcolor{gray}{36.1} & \textcolor{gray}{76.3} & \textcolor{gray}{25.7} & \textcolor{gray}{82.8} & \textcolor{gray}{35.6} & \textcolor{gray}{0.0} & \textcolor{gray}{0.0} \\
    \multicolumn{2}{c|}{\textbf{Ours} (GPT 3.5+Diverse+Max)} & 77.7 & 19.4 & 78.5 & 19.1 & 82.1 & 34.2 & 84.7 & 37.7 & 79.9 & 28.1 & 83.9 & 36.8 & +3.0 & +2.9 \\
    \midrule[0.3pt]
    \multirow{4}{*}{\textbf{Model}}       & PEGASUS        & $\times$ & 18.2 & $\times$ & 18.5 & $\times$ & 33.2 & $\times$ & 37.0 & $\times$ & 27.4 & $\times$ & 36.0 & $\times$ & +2.0 \\
                                          & Parrot         & $\times$ & 17.5 & $\times$ & 18.3 & $\times$ & 32.2 & $\times$ & 36.8 & $\times$ & 26.3 & $\times$ & 36.1 & $\times$ & +1.5 \\
                                          & QCPG           & $\times$ & 17.4 & $\times$ & 17.2 & $\times$ & 32.8 & $\times$ & 37.0 & $\times$ & 26.8 & $\times$ & 36.2 & $\times$ & +1.5 \\
                                          & LLaMA-2-70b-chat & 74.5 & 17.5 & 76.3 & 16.6 & 79.2 & 32.9 & 83.6 & 36.8 & 78.8 & 26.8 & 82.5 & 36.3 & +1.1 & +1.4 \\
    \midrule[0.3pt]
    \multirow{2}{*}{\textbf{Prompt}}      & Basic          & 77.4 & 17.6 & 77.4 & 16.9 & 81.8 & 33.2 & 83.9 & 37.1 & 79.2 & 27.1 & 83.2 & 36.3 & +2.4 & +1.7 \\
                                          & Multilingual   & 77.7 & --   & 77.7 & --   & 81.8 & --   & 84.7 & --   & 79.2 & --   & 83.9 & --   & +2.7 & 0.0   \\
    \midrule[0.3pt]
    \multirow{2}{*}{\textbf{Aggregation}} & Mean           & 77.0 & 16.6 & 78.8 & 10.5 & 83.2 & 32.2 & 81.8 & 35.5 & 79.2 & 23.1 & 81.8 & 33.9 & +2.2 & -1.1 \\
                                          & Built-in       & 78.5 & 18.8 & 78.5 & 19.1 & $\times$ & $\times$ & $\times$ & $\times$ & $\times$ & $\times$ & $\times$ & $\times$ & $\times$ & $\times$   \\
    \midrule[0.3pt]
    \multicolumn{2}{c|}{\textbf{Filtering subpar references}} & 77.7 & 19.2 & 78.5 & 19.0 & 82.1 & 34.1 & 84.3 & 37.6 & 79.9 & 28.0 & 83.9 & 36.8 & 0.0 & -0.1      \\
    \bottomrule
    \end{tabular}}
    \caption{Analysis of the effect of the diversifying models, instruction prompts, aggregation functions, and post-filtering. We report the system-level accuracy and segment-level correlation of the Chinese-to-English direction over the WMT22 Metric Task. $\times$ of PEGASUS, Parrot, and QCPG denotes the three methods do not support multilingual scenario. $\times$ of ``Bulit-in'' means the metric do not have built-in multi-reference aggregation option. --~in ``Multilingual'' represents the multilingual diverse prompt has the same results as the English diverse prompt.}
    \label{tab:my_ablation}
\end{table*}

\begin{table*}[t]
    \centering
    \resizebox{1\textwidth}{!}{
    \begin{tabular}{ll|cc|cc|cc|cc|cc|cc|cc}
    \toprule
    \multicolumn{2}{c|}{\multirow{2}{*}{\textbf{Settings}}}                             & \multicolumn{2}{c|}{\textbf{BLEU}} & \multicolumn{2}{c|}{\textbf{ChrF}}  & \multicolumn{2}{c|}{\textbf{BERTScore}} & \multicolumn{2}{c|}{\textbf{BLEURT}} & \multicolumn{2}{c|}{\textbf{Prism}} & \multicolumn{2}{c|}{\textbf{COMET}} & \multicolumn{2}{c}{\textbf{Average Gains}} \\ 
    & & En-De & En-Ru & En-De & En-Ru & En-De & En-Ru & En-De & En-Ru & En-De & En-Ru & En-De & En-Ru & En-De & En-Ru \\
    \midrule
    \multicolumn{2}{c|}{\textbf{\textcolor{gray}{Single-Ref}}}                 & \textcolor{gray}{16.9} & \textcolor{gray}{14.0} & \textcolor{gray}{21.4} & \textcolor{gray}{16.8} & \textcolor{gray}{23.2} & \textcolor{gray}{19.2} & \textcolor{gray}{34.4} & \textcolor{gray}{35.9} & \textcolor{gray}{21.5} & \textcolor{gray}{23.0} & \textcolor{gray}{34.3} & \textcolor{gray}{37.2} & \textcolor{gray}{0.0} & \textcolor{gray}{0.0}  \\
    \multicolumn{2}{c|}{\textbf{Ours} (GPT 3.5+Diverse+Max)} & 19.3 & 17.9 & 24.5 & 21.6 & 25.9 & 23.5 & 34.7 & 37.1 & 23.4 & 26.1 & 35.0 & 38.5 & +1.9 & +3.1  \\
    \midrule[0.3pt]
    \textbf{Model}                        & LLaMA-2-70b-chat & 18.1 & 16.0 & 22.8 & 19.5 & 24.1 & 21.6 & 34.8 & 36.8 & 22.4 & 24.7 & 35.1 & 38.2 & +0.9 & +1.7 \\
    \midrule[0.3pt]
    \multirow{2}{*}{\textbf{Prompt}}      & Basic            & 19.6 & 19.3 & 25.2 & 24.2 & 26.2 & 25.4 & 35.5 & 34.7 & 23.9 & 23.0 & 35.2 & 34.8 & +2.3 & +2.6      \\
                                          & Multilingual     & 18.9 & 19.1 & 22.4 & 22.2 & 23.9 & 24.2 & 37.3 & 37.1 & 26.4 & 26.1 & 38.7 & 38.9 & +2.7 & +3.6        \\
    \midrule[0.3pt]
    \multirow{2}{*}{\textbf{Aggregation}} & Mean             & 13.9 & 15.0 & 17.2 & 16.3 & 20.0 & 19.4 & 32.3 & 37.0 & 19.2 & 22.3 & 32.0 & 36.6 & -2.8 & +0.1      \\
                                          & Built-in         & 18.4 & 18.1 & 24.5 & 21.6 & $\times$ & $\times$ & $\times$ & $\times$ & $\times$ & $\times$ & $\times$ & $\times$ & $\times$ & $\times$  \\
    \midrule[0.3pt]
    \multicolumn{2}{c|}{\textbf{Filtering subpar references}}& 19.4 & 17.9 & 24.8 & 21.6 & 26.0 & 23.5 & 34.8 & 37.1 & 23.4 & 26.1 & 35.1 & 38.5 & +0.2 & 0.0      \\
    \bottomrule
    \end{tabular}}
    \caption{Ablation analysis in the English-to-German and English-to-Russia and directions using segment-level Kendall Tau correlation.}
    \label{tab:my_ablation2}
\end{table*}

\begin{table*}[t]
    \centering
    \resizebox{1\textwidth}{!}{
    \begin{tabular}{c|ccccccc}
    \toprule
    \textbf{Prompts} & \textbf{BLEU} & \textbf{ChrF} & \textbf{BERTScore} & \textbf{BLEURT} & \textbf{Prism} & \textbf{COMET} & \textbf{Average Gains} \\
    \midrule
    \textcolor{gray}{\textbf{Single-Ref}} & \textcolor{gray}{14.5} & \textcolor{gray}{14.7} & \textcolor{gray}{31.6} & \textcolor{gray}{36.1} & \textcolor{gray}{25.7} & \textcolor{gray}{35.6} & \textcolor{gray}{0.0} \\
    \textbf{Ours (Mixing \ding{192}-\ding{201})} & 19.4 & 19.1 & 34.2 & 37.7 & 28.1 & 36.8 & +2.9 \\
    \midrule[0.3pt]
    \ding{192} $\times$ 10 & 16.6 & 16.3 & 33.0 & 37.1 & 26.8 & 36.3 & +1.3 \\
    \ding{193} $\times$ 10 & 15.9 & 15.5 & 32.2 & 36.4 & 26.4 & 35.7 & +0.6 \\
    \ding{194} $\times$ 10 & 17.8 & 17.5 & 33.0 & 36.8 & 27.0 & 36.2 & +1.7 \\
    \ding{195} $\times$ 10 & 16.8 & 16.7 & 32.8 & 36.9 & 26.6 & 36.0 & +1.3 \\
    \ding{196} $\times$ 10 & 15.1 & 15.4 & 32.0 & 36.3 & 26.1 & 35.6 & +0.4 \\
    \ding{197} $\times$ 10 & 18.1 & 17.5 & 33.5 & 37.4 & 27.4 & 36.3 & +2.0 \\
    \ding{198} $\times$ 10 & 17.4 & 16.5 & 33.4 & 37.2 & 27.0 & 36.4 & +1.6 \\
    \ding{199} $\times$ 10 & 18.1 & 17.2 & 33.4 & 37.4 & 27.2 & 36.4 & +1.9 \\
    \ding{200} $\times$ 10 & 16.8 & 16.2 & 33.1 & 37.3 & 26.8 & 36.2 & +1.4 \\
    \ding{201} $\times$ 10 & 18.6 & 19.0 & 33.7 & 37.2 & 27.5 & 36.5 & +2.4 \\
    \bottomrule
    \end{tabular}}
    \caption{Diverse prompts analysis in the Chinese-to-English direction using segment-level Kendall Tau correlation.}
    \label{tab:my_ablation3}
\end{table*}

\begin{figure}[t]
    \centering
    \includegraphics[width=1.0\columnwidth]{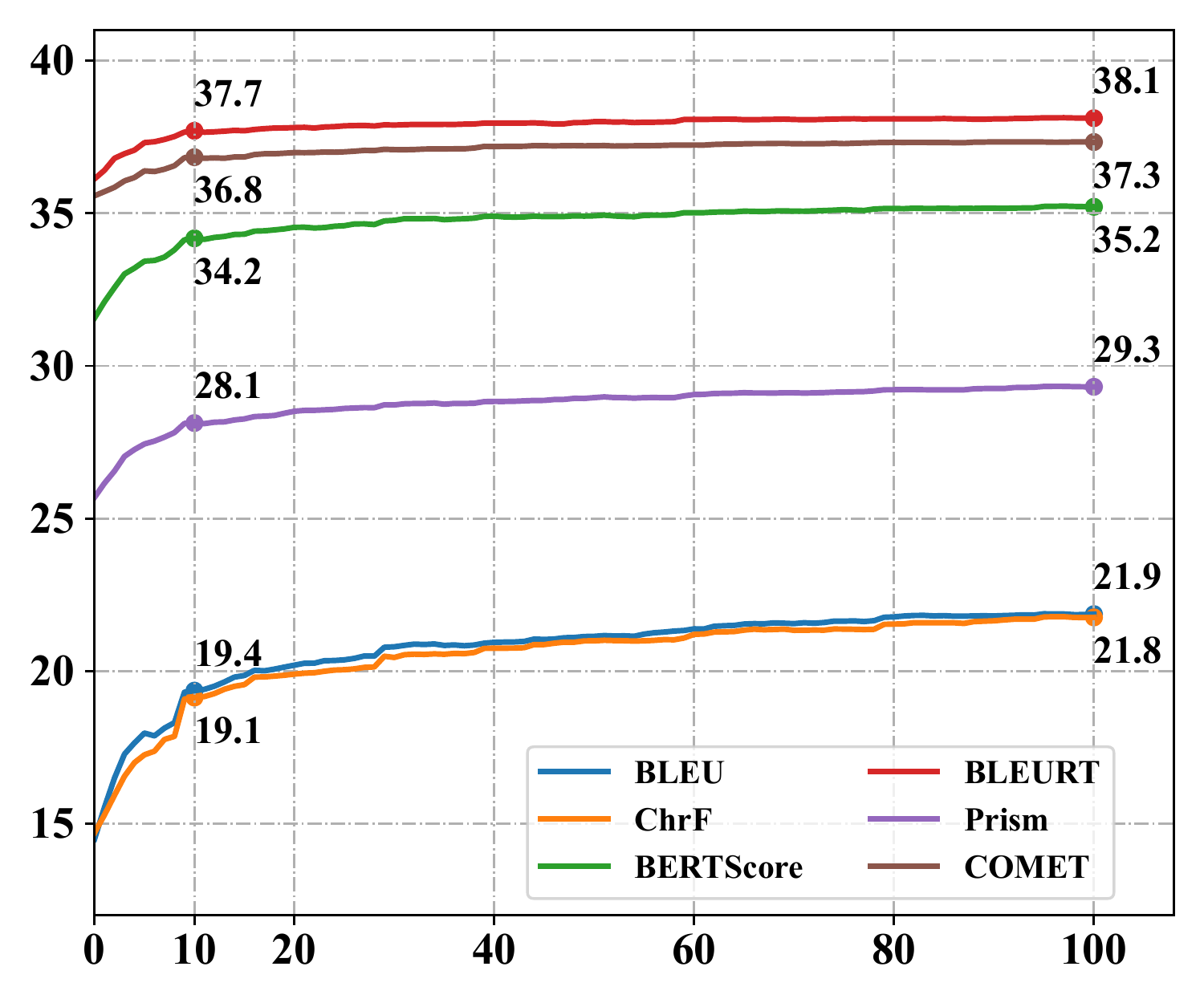}
    \caption{Kendall Tau correlation score \wrt the number of generated references in the Chinese-to-English direction on the WMT22 Metrics Shared Task.}
    \label{fig:num-Ref}
\end{figure}

In this section, we examine the impact of various factors of increasing the reference numbers, which include the selection of diversifying models, the application of instruction prompts, the choice of the aggregation function, the effect of post-filtering, and the number of diversified references. The results can be found in Table~\ref{tab:my_ablation} and~\ref{tab:my_ablation2} and Figure~\ref{fig:num-Ref}.

(1) Firstly, we compare the influence of our diversifying LLM \texttt{gpt-3.5-turbo-instruct} with three rephrasing PLMs PEGASUS-Paraphrasing\footnote{\url{https://huggingface.co/tuner007/pegasus_paraphrase}}, Parrot\footnote{\url{https://huggingface.co/prithivida/parrot_paraphraser_on_T5}}, and QCPG~\cite{bandel-etal-2022-quality}, which are fine-tuned on paraphrasing tasks. However, these three models only support English paraphrasing. We also incorporate another open-source LLMs, LLaMA-2-70b-chat, to diversify our references. From the results, we observe that \texttt{gpt-3.5-turbo-instruct} can outperform three PLMs and LLaMA-2-chat in all metrics, which showcases its superior capability in completing the semantic space of given reference.

(2) Regarding the choice of instruction prompts, we first degrades the diverse prompts to the basic prompt mentioned in Section~\ref{sec:para}. We observe that the diverse prompts can achieve satisfactory results on English references (\ie Zh-En), and may slightly reduce the performance on non-English languages (Table~\ref{tab:my_ablation2}). Then, we further translate the English diverse prompts into respective language (\ie instructing LLMs using the reference language), and find the gains of multilingual diverse prompts are also not obvious. We attribute the two results to that fact the diversifying ability of LLMs in non-English is not as good as that in English, since English is the dominant language. 
Besides, we analyze each kind of our diverse prompts in Appendix. We compare a mixture of one sentence per prompt with ten sentences per prompt. From the results in Table~\ref{tab:my_ablation3}, we can find that mixing prompts is better than any individual prompt. This further demonstrates the effectiveness of our delicate prompts and they can cover a broader semantics range of reference sentences.

(3) Thirdly, we investigate the aggregation functions using the mean aggregation and the built-in multi-reference aggregation of BLEU and ChrF. We discover that when changing the aggregation from \textit{maximum} to \textit{mean}, the correlation scores for most metrics have dropped, especially in the Chinese-to-English direction. This indicates that the highest-quality reference plays a dominant role in generation evaluation, and our approach to increasing the number of references significantly strengthens this probability. However, averaging multiple reference scores could introduce noise from low-quality reference scores. As for the built-in method of BLEU and ChrF, their performances are indistinguishable.

(4) In addition, we attempt to filter the generated references considering some of them may be of low quality. We employ \texttt{gpt-3.5-turbo} to judge using the instruction: ``\textit{Sentence 1: \{ref\}\textbackslash nSentence 2: \{div\_ref\}\textbackslash nDo sentence 1 and sentence 2 convey the same meaning?\textbackslash n\textbackslash n}''. After eliminating the reference unrecognized by \texttt{gpt-3.5-turbo}, we can find that the removal of low-quality sentences has minimal impact on correlation results. We speculate that our approach involves aggregating results from multiple references and selecting the one with the highest score, effectively disregarding those of inferior quality.

(5) Finally, we examine the influence of scaling the number of references. We utilize the diverse prompts to generate more references. From Figure~\ref{fig:num-Ref}, we observe a consistent upward trend in the overall performance as the number of references increases. 
For word-based metrics, this growth trend is more obvious. 
This experiment further shows that traditional benchmarks that relies on a single reference is very one-sided for NLG evaluation, and we need to provide multiple references for benchmarks.
Considering that the performance of neural metrics tends to saturate when the quantity is high, over-generation may not lead to more significant gains, suggesting that the optimal cost-effective number may not exceed 20. 

\section{Conclusion}
In this paper, we have investigated the effect of enriching the number of references in NLG benchmarks and verified its effectiveness.
Our diversifying method, Div-Ref, can effectively cover the semantic space of the golden reference, which can largely extend the limited references in existing benchmarks.
With extensive experiments, our approach yields substantial improvements in the consistencies  between evaluation metrics and human evaluation.
In future work, we will explore the current evaluation method on more NLG tasks, and also consider extending it to evaluate generation tasks in other modalities. It is also valuable to investigate whether paraphrasing can improve LLMs' training and utilization.


\section*{Acknowledgement}
This work was partially supported by Beijing Natural Science Foundation under Grant No. L233008 and 4222027. Xin Zhao is the corresponding author.

\section*{Limitations}
Despite conducting numerous experiments, further research is required to explore the number of references and the optimal diversifying techniques that can achieve a trade-off between time and effectiveness. Since using more references leads to more evaluation time, future work can explore strategies for mitigating these issues, possibly through the implementation of a selection mechanism that prioritizes sentences with diverse expressions while minimizing the overall number of reference sentences.
Moreover, Our diverse prompts may fail in specialized domains, such as finance and biomedicine. Rewriting professional terms may lead to inaccuracy evaluation of the generated sentences. Future work can further investigate and validate the effectiveness of our method within these domains. Additionally, we can design more fine-grained prompts tailored to address the specific challenges posed by professional terminology.
In addition, due to the high cost of \texttt{text-davinci-003}, we omit the experiments of GEMBA in the ablation analysis, which may lead to an incomplete analysis of LLM-based metrics. The OpenAI API also is non-deterministic, which may lead to different diversifying results for the same input. There is also a chance that OpenAI will remove existing models.


\bibliography{ref}

\appendix
\newpage

\section{Experimental Details}
\subsection{Metric Implementation} \label{app:metric}
The implementation details of each metric in different benchmarks are listed as follows:
\begin{itemize}[leftmargin=*]
\itemsep0em 
\item ChrF~\cite{chrf}: We utilize sentence-level ChrF from \texttt{SacreBLEU}\footnote{\url{https://github.com/mjpost/sacrebleu}} for machine translation.
\item BLEU~\cite{bleu}: We utilize sentence-level BLEU from \texttt{SacreBLEU}\footnote{\url{https://github.com/mjpost/sacrebleu}} for machine translation, and employ BLEU from \texttt{pycocoevalcap}\footnote{\url{https://github.com/salaniz/pycocoevalcap}} for image caption.
\item ROUGE-1/2/L~\cite{rouge}: We utilize ROUGE-1/2/L from \texttt{files2rouge}\footnote{\url{https://github.com/pltrdy/files2rouge}} for text summarization, and employ ROUGE-L from \texttt{pycocoevalcap}\footnote{\label{foot:coco}\url{https://github.com/salaniz/pycocoevalcap}} for image caption.
\item METEOR~\cite{meteor}: We utilize METEOR from \texttt{pycocoevalcap}\footref{foot:coco} for image caption.
\item CIDEr~\cite{meteor}: We utilize CIDEr from \texttt{pycocoevalcap}\footref{foot:coco} for image caption.
\item SPICE~\cite{meteor}: We utilize SPICE from \texttt{pycocoevalcap}\footref{foot:coco} for image caption.
\item BERTScore~\cite{bertscore}: We utilize BERTScore from its official repository\footnote{\url{https://github.com/Tiiiger/bert_score}} for machine translation, text summarization, and image caption. Specially, we leverage \texttt{roberta-large} for English reference sentences, while apply \texttt{bert-base-multilingual-cased} for other languages (\ie German and Russia).
\item MoverScore~\cite{moverscore}: We utilize MoverScore from its official repository\footnote{\url{https://github.com/AIPHES/emnlp19-moverscore}} for text summarization. Specially, we leverage the \texttt{MNLI-BERT} checkpoint.
\item BLEURT~\cite{bleurt}: We utilize BLEURT from its official repository\footnote{\url{https://github.com/google-research/bleurt}} for machine translation. Specially, we leverage the \texttt{BLEURT-20} checkpoint.
\item Prism~\cite{prism}: We utilize Prism from its official repository\footnote{\url{https://github.com/thompsonb/prism}} for machine translation.
\item COMET~\cite{comet}: We utilize COMET from its official repository\footnote{\url{https://github.com/Unbabel/COMET}} for machine translation. Specially, we leverage the \texttt{Unbabel/wmt22-comet-da} checkpoint.
\item BARTScore~\cite{bartscore}: We utilize BARTScore from its official repository\footnote{\url{https://github.com/neulab/BARTScore}} for machine translation in the Chinese-to-English direction. Specially, we leverage the \texttt{BARTScore+CNN+Para} checkpoint.
\item GEMBA~\cite{gemba}: We utilize GEMBA-Dav3-DA from its official repository\footnote{\url{https://github.com/MicrosoftTranslator/GEMBA}} for machine translation. Specially, we leverage direct assessment as the scoring task, and apply \texttt{text-davinci-003} as the evaluation model with temperature=0. 
\item ChatGPT-eval~\cite{wang2023chatgpt}: We utilize ChatGPT-eval (Stars w/ ref) from its official repository\footnote{\url{https://github.com/krystalan/chatgpt_as_nlg_evaluator}} for text summarization. Specially, we leverage the star prompt with reference, and apply \texttt{gpt-3.5-turbo} as the evaluation model with temperature=0. 
\end{itemize}


\subsection{Diversified Examples}

\begin{table*}
    \centering
    \begin{tabular}{p{0.5\textwidth}}
        \toprule
        \textbf{Source} \\
        \begin{CJK*}{UTF8}{gbsn}是否有途径处罚他\end{CJK*} \\
        \midrule
        \textbf{Ground-truth reference} \\
        Is there a way to punish him? \\
        \midrule
        \textbf{Diversified references} \\
        Can he be penalized? \\
        Is there a way to punish him? \\
        Can he be punished in any way? \\
        Can he be punished? \\
        Was there a way to punish him? \\
        Can he be punished in any way? \\
        Is there a means of reprimanding him? \\
        Can he be punished in any way? \\
        Is there a means by which he may be disciplined? \\
        Can we do something to punish him? \\
        \bottomrule
    \end{tabular}
    \caption{The diversified example of WMT22 Metrics  Task in the Chinese-to-English direction. More examples can be found at \url{https://github.com/RUCAIBox/Div-Ref}.}
    \label{tab-zh}
\end{table*}

\begin{table*}
    \centering
    \begin{tabular}{p{0.7\textwidth}}
        \toprule
        \textbf{Source} \\
        I sincerely hope you get to find a resolution \\
        \midrule
        \textbf{Ground-truth reference} \\
        Ich hoffe wirklich, dass Sie eine Lösung finden werden \\
        \midrule
        \textbf{Diversified references} \\
        Ich drücke die Daumen, dass Sie eine Lösung ausarbeiten können. \\
        Ich hoffe wirklich, dass Sie eine Lösung finden werden. \\
        Ich hoffe, dass Sie eine Lösung finden werden, wirklich. \\
        Ich hoffe wirklich, dass eine Lösung von Ihnen gefunden werden wird. \\
        Ich hatte gehofft, dass Sie eine Lösung finden würden. \\
        Ich hoffe, dass Sie eine Lösung finden werden \\
        Ich wünsche Ihnen aufrichtig, dass Sie eine Lösung finden werden. \\
        Ich wünsche mir innigst, dass Sie eine Lösung finden werden. \\
        Ich hoffe aufrichtig, dass Sie eine Lösung finden werden. \\
        Ich hoffe wirklich, dass du eine Lösung findest. \\
        \bottomrule
    \end{tabular}
    \caption{The diversified example of WMT22 Metrics  Task in the English-to-German direction. More examples can be found at \url{https://github.com/RUCAIBox/Div-Ref}.}
    \label{tab-de}
\end{table*}

\begin{table*}
    \centering
    \begin{tabular}{p{0.95\textwidth}}
        \toprule
        \textbf{Source} \\
        I see it all the time in my line of work. \\
        \midrule
        \textbf{Ground-truth reference} \\
        \foreignlanguage{russian}{Я постоянно вижу такое в своей сфере деятельности.} \\
        \midrule
        \textbf{Diversified references} \\
        \foreignlanguage{russian}{Я всегда наблюдаю за подобным в своей сфере работы.} \\
        \foreignlanguage{russian}{Такое я вижу постоянно в своей сфере деятельности.} \\
        \foreignlanguage{russian}{Такое я постоянно вижу в своей сфере деятельности.} \\
        \foreignlanguage{russian}{Такое постоянно видится мной в моей сфере деятельности.} \\
        \foreignlanguage{russian}{Я постоянно увижу такое в своей сфере деятельности.} \\
        \foreignlanguage{russian}{В своей сфере деятельности я часто наблюдаю подобное.} \\
        \foreignlanguage{russian}{Я всегда наблюдаю подобное в своей сфере работы.} \\
        \foreignlanguage{russian}{В своей сфере деятельности я непрерывно наблюдаю подобное.} \\
        \foreignlanguage{russian}{Я постоянно наблюдаю подобные вещи в своей сфере профессиональной деятельности.} \\
        \foreignlanguage{russian}{Я всегда это наблюдаю в своей работе.} \\
        \bottomrule
    \end{tabular}
    \caption{The diversified example of WMT22 Metrics  Task in the English-to-Russian direction. More examples can be found at \url{https://github.com/RUCAIBox/Div-Ref}.}
    \label{tab-ru}
\end{table*}

\end{document}